\documentclass[10pt,twocolumn,letterpaper]{article}

\usepackage{cvpr}
\usepackage{times}
\usepackage{epsfig}
\usepackage{graphicx}
\usepackage{amsmath}
\usepackage{csquotes}
\usepackage{multicol}
\usepackage{framed,multirow}
\usepackage{amssymb}
\usepackage{latexsym}
\usepackage{url}
\usepackage{xcolor}
\definecolor{newcolor}{rgb}{.8,.349,.1}
\usepackage{epsfig}
\usepackage{amsmath,amssymb} 
\usepackage[export]{adjustbox}
\usepackage{array}
\usepackage{rotating}
\usepackage{booktabs} 
\usepackage{multirow}
\usepackage{color}
\usepackage{color, colortbl}
\usepackage{float}
\usepackage{xspace}

\usepackage[pagebackref=true,breaklinks=true,letterpaper=true,colorlinks,bookmarks=false]{hyperref}

\cvprfinalcopy 

\def\cvprPaperID{0023} 

\begin{document}

\title{Identity Enhanced Residual Image Denoising}

\author{Saeed Anwar$^{1,2}$, Cong Phuoc Huynh$^2$, Fatih Porikli$^2$\\
$^1$Data61-CSIRO,\quad $^2$The Australian Natinal University\\
{\tt\small saeed.anwar@data61.csiro.au}
}

\maketitle
\thispagestyle{empty}

\begin{abstract}
We propose to learn a fully-convolutional network model that consists of a Chain of Identity Mapping Modules and residual on the residual architecture for image denoising. Our network structure possesses three distinctive features that are important for the noise removal task. Firstly, each unit employs identity mappings as the skip connections and receives pre-activated input to preserve the gradient magnitude propagated in both the forward and backward directions. Secondly, by utilizing dilated kernels for the convolution layers in the residual branch, each neuron in the last convolution layer of each module can observe the full receptive field of the first layer.  Lastly, we employ the residual on the residual architecture to ease the propagation of the high-level information. Contrary to current state-of-the-art real denoising networks, we also present a straightforward and single-stage network for real image denoising.

The proposed network produces remarkably higher numerical accuracy and better visual image quality than the classical state-of-the-art and CNN algorithms when being evaluated on the three conventional benchmark and three real-world datasets.
\end{abstract}

\section{Introduction}
\begin{figure}[t!bp]
\begin{center}
\begin{tabular}{c@{ } c@{ } c}  
\includegraphics[trim={2.5cm 5.5cm  2.5cm  4cm },clip,width=.15\textwidth]{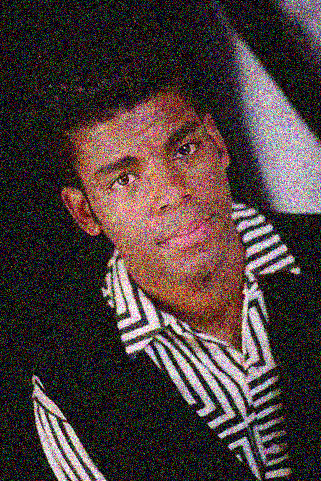}&  
\includegraphics[trim={2.5cm 5.5cm  2.5cm  4cm },clip,width=.15\textwidth]{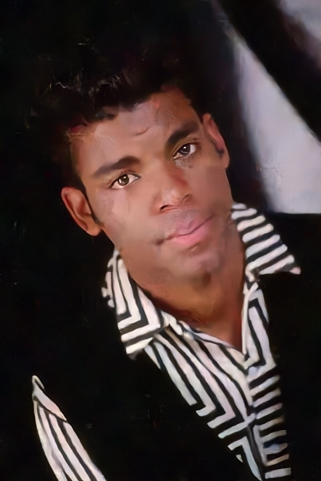}&
\includegraphics[trim={2.5cm 5.5cm  2.5cm  4cm },clip,width=.15\textwidth]{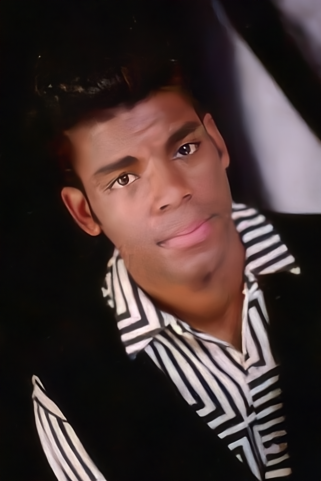}\\
14.16dB &  32.05dB &  32.64dB\\
Input   &   DnCNN &  Proposed \\

\includegraphics[trim={5cm 2cm  5cm  2cm},clip,width=.15\textwidth]{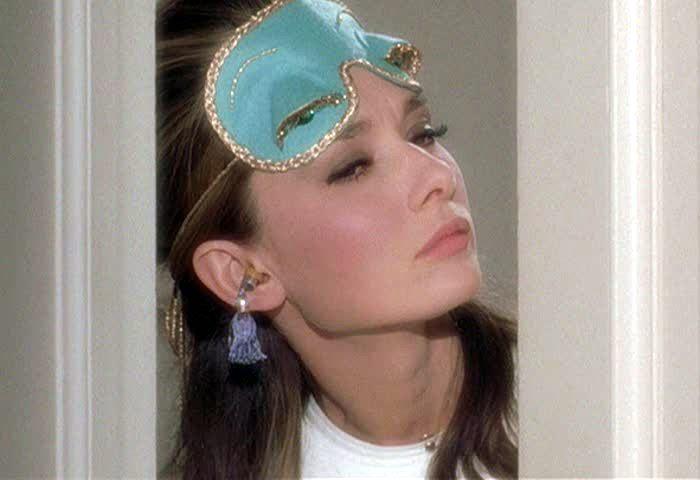}&  
\includegraphics[trim={5cm 2cm  5cm  2cm},clip,width=.15\textwidth]{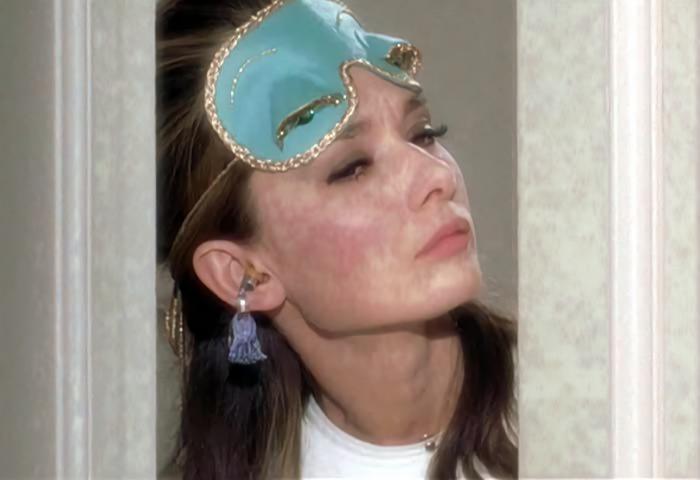}&
\includegraphics[trim={5cm 2cm  5cm  2cm},clip,width=.15\textwidth]{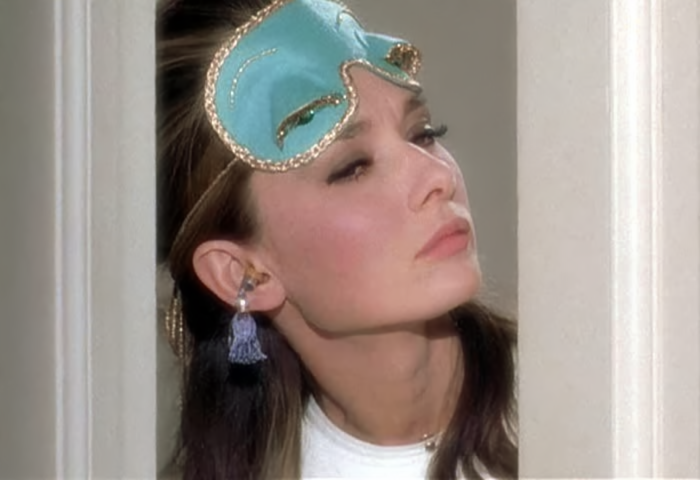}\\
Input   &   CBDNet &  Proposed \\

\end{tabular}
\end{center}
\vspace*{-1mm}
\caption {Denoising results: In the first row, an image corrupted by the Gaussian noise with $\sigma=50$ from the BSD68 dataset~(\cite{roth2009fields}). In the second row, a sample image from RNI15~(\cite{lebrun2015NC}) real noisy dataset. Our results have the best PSNR score for synthetic images, and unlike other methods, it does not have over-smoothing or over-contrasting artifacts. Best viewed in color on a high-resolution display.}
\label{fig:sample_front_image}
\vspace*{-3mm}
\end{figure}

In recent years, the amount of multimedia content is growing at an enormous rate, for example, online videos, audios, and photos due to hand-held devices and other types of multimedia devices. Thus, image processing, specifically, image denoising, has become an essential process for various computer vision and image analysis applications.  A few notable methods benefiting from image denoising are detection~(\cite{rozantsev2015rendering}), face recognition~(\cite{hjelmaas2001face}), super-resolution~(\cite{yang2018video}), \etc  In the past few years, the research in this area has shifted its focus on how to make the best use of image priors. To this end, several approaches attempted to exploit non-local self similar (NSS) patterns~(\cite{Buades2005NLM,Dabov2007BM3D}), sparse models~(\cite{Gu2014WNN,peng2012rasl}), gradient models~(\cite{xu2007Iterative,weiss2007makes}), Markov random field models~(\cite{roth2009fields}), external denoising~(\cite{Yue2014CID,anwar2017category,luo2015adaptive}) and convolutional neural networks~(\cite{zhang2017DnCNN, lefkimmiatis2017NLNet, zhang2017IRCNN}).\let\thefootnote\relax\footnotetext{Code available at https://github.com/saeed-anwar/IERD}

The non-local matching (NLM) of self-similar patches and block matching with 3D filtering (BM3D) in a collaborative manner have been two prominent baselines for image denoising for almost a decade now. Due to popularity of NLM~(\cite{Buades2005NLM}) and BM3D~(\cite{Dabov2007BM3D}), a number of their variants~(\cite{Foi2007SADCT,Lebrun2013NLB,Goossens2008INLM}) were also proposed to execute the search for similar patches in similar transform domains.

 The use of external priors for denoising has been motivated by the pioneering studies of ~\cite{Levin2011Bounds,Chatterjee2010IDD}, which showed that selecting correct reference patches from a large external image dataset of clean samples can theoretically suppress additive noise and attain infinitesimal reconstruction error. However, directly incorporating patches from an external database is computationally prohibitive even for a single image. To overcome this problem, Chan~\etal~\cite{chan2014monte} proposed efficient sampling techniques for large databases but still the denoising is impractical as it takes hours to search patches for one single image if not days. An alternative to these methods can be considered as the dictionary learning based approaches~\cite{Elad2009ERD,Mairal2009NLSM,Dong2011CSR}, which learn over-complete dictionaries from a set of external natural clean images and then enforce patch self-similarity through sparsity. 

Aiming at improving the use of external datasets, many previous works such as~\cite{Zoran2011EPLL,Chen2015External,Xu2015PG-GMM} investigated the use of maximum likelihood frameworks to learn Gaussian mixture models of natural image patches or group patches for clean patch estimation. Several studies, including ~\cite{Xu2015PGPD,chen2015PCLR}, modified Zoran~\etal~\cite{Zoran2011EPLL}'s statistical prior for reconstruction of class-specific noisy images by capturing the statistics of noise-free patches from a large database of same category images through the Expectation-Maximization algorithm. Other similar methods on external denoising include TID~\cite{luo2015adaptive}, CSID~\cite{anwar2017category} and CID~\cite{Yue2015CID}; however, all of these have limited applicability in denoising of generic (from an unspecific class) images.

As an alternative, CSF \cite{schmidt2014CSF} learns a single framework based on unification of random-field based model and half-quadratic optimization. The role of the shrinkage in wavelet image restoration is to attenuate small values towards zero due to the assumption of these values being the product of noise instead of the signal values.These predictions are then chained to form a cascade of shrinkage fields of Gaussian conditional random Fields. The CSF algorithm considers the data term to be quadratic and must have a closed-form solution based on discrete Fourier transform.

With the rise of convolutional neural networks (CNN), a significant performance boost for image denoising has been achieved \cite{zhang2017DnCNN,zhang2017IRCNN,lefkimmiatis2017NLNet,Burger2012MLP,schmidt2014CSF}. Using deep neural networks, IrCNN~\cite{zhang2017IRCNN} and DnCNN~\cite{zhang2017DnCNN} learn to predict the residual noise present in the contaminated image by using the ground-truth noise in the loss function instead of the clean image. The architectures of IrCNN \cite{zhang2017IRCNN} and DnCNN \cite{zhang2017DnCNN} are very simple as it only stacks of convolutional, batch normalization and ReLU layers. Although both models were able to report favorable results, their performance depends heavily on the accuracy of noise estimation without knowing the underlying structures and textures present in the image. 

TRND~\cite{chen2017TNRD} incorporated a field-of-experts prior~\cite{roth2009fields} into its convolutional network by extending conventional nonlinear diffusion model to highly trainable parametrized linear filters and influence functions. It has shown improved results over more classical methods; however, the imposed image priors inherently impede its performance, which highly rely on the choice of hyper-parameter settings, extensive fine-tuning and stage-wise training.

 Another notable deep learning-based work is non-local color image denoising (abbreviated as NLNet), presented by \cite{lefkimmiatis2017NLNet} which exploits the non-local self-similarity using deep networks. Non-local variational schemes have motivated the design of the NLNet model \cite{lefkimmiatis2017NLNet} and employ the non-local self-similarity property of natural images for denoising.  The performance heavily depends on coupling discriminative learning and self-similarity. The restoration performance is comparatively better to several earlier state-of-the-art. Though, this model improves on classical methods but lagging behind  IrCNN~\cite{zhang2017IRCNN} and DnCNN~\cite{zhang2017DnCNN}, as it inherits the limitations associated with the NSS priors as not all patches recur in an image.

Currently, the trend changed from synthetic denoising towards real-image denoising~(\cite{plotz2018N3Net,guo2018CBDnet,brooks2019UPI,anwar2019real}). Although, the algorithms, for example, DnCNN, \etc trained a single model for synthetic datasets; however, it failed to achieve satisfactory results on real images. Commonly, real-image denoising is a two-stage process. The first step involves the prediction of the noise variance, while the second stage employs the predicted noise-level to denoise the image. As an example,  Noise Clinic proposed (NC)  by \cite{lebrun2015NC} first predicts the noise, which is dependent on the signal's frequency and then used non-local Bayes (NLB)~(\cite{Lebrun2013NLB}) to denoise it.

Similarly,~\cite{zhang2018ffdnet} trains FFDNet, a non-blind denoising network based on Gaussian noise. The mentioned network achieves partial success in denoising the real noisy images. However, FFDNet requires manual settings in case of high noise variance.  More recently,~\cite{guo2018CBDnet} proposed CBDNet, a blind network for real-noisy images.  The system is composed of two subnets: one for prediction of noise and the second to denoise photographs using the predicted noise. Furthermore, CBDNet uses multiple losses and exploits synthetic and real images alternatively to train the model. The authors also report the use of high noise variance to denoise a low noisy image. Moreover, to improve results, the system may require manual intervention. More recently, Anwar \& Barnes presented denoising real images via attention mechanism, known as RIDNet~\cite{anwar2019real}. The modules are carefully designed to learn features differently. In this work we present a straightforward end-to-end structure that delivers results on real noisy images using a single-stage network without requiring any intervention or attention mechanism.

\subsection{Inspiration \& Motivation}

\begin{figure*}
\begin{center}
\includegraphics[trim={6.3cm 7.4cm 4.5cm 4.5cm},clip,width=\textwidth]{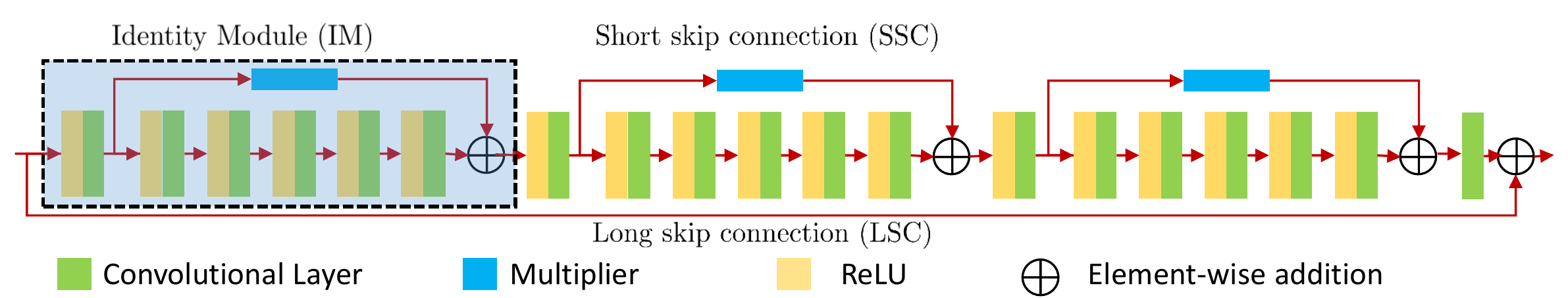}\\ 
\end{center}
\caption {The proposed network architecture, which consists of multiple modules with similar structures. Each module is composed of a series of pre-activation-convolution layer pairs. The multiplier block negates the input block features to be summed at the end of the mapping module.}
\label{fig:net_architecture}
\end{figure*}

Existing convolutional neural network image denoising methods~(\cite{Burger2012MLP,zhang2017DnCNN,zhang2017IRCNN}) connect weight layers consecutively and learn the mapping by brute force. One problem with such an architecture is the addition of more weight layers to increase the depth of the network. Even if the new weight layers are added to the above mentioned CNN based denoising methods, it will suffer from the vanishing  gradients problem and make it worse~(\cite{bengio1994vanishing}). This increase in the depth of the network is essential to attain the performance boost~(\cite{he2016deep}). Therefore, our goal is to propose a model that overcomes this deficiency.  Another reason is the lack of single-stage real image denoising. Most of the current denoising systems are either for synthetic image denoising or treat noise estimation and denoising separately, ignoring the relationship between the noise and the image structures.

To provide a solution, our choice is the convolutional neural networks in a discriminative prior setting for image denoising. There are many advantages of using single-stage CNNs for synthetic and real images, which include efficient inference,  incorporation of robust priors, integration of local and global receptive fields, regressing on nonlinear models, and discriminative learning capability. Furthermore, we propose a modular single-stage network where we call each module as a identity module (IM). The identity module can be replicated and easily extended to any arbitrary depth for performance enhancement.

\subsection{Contributions}
The contributions of this work can be summarized as follows:
\begin{itemize}
\item An effective CNN architecture that consists of a Chain of Identity Mapping modules for image denoising. These modules share a common composition of layers, with residual connections between them to facilitate training stability.
\item The use of dilated convolutions for learning suitable filters to denoise at different levels of spatial extent and residual on the residual architecture for the ease of flow of the high-frequency details.
\item A low-weight single-stage real image denoiser without any complex modules.
\item Extensive evaluation on six datasets (three synthetic and three real) against more than 20 state-of-the-art denoising methods.
\end{itemize}

\section{Identity Enhanced Residual Denoising}
\label{sec:IERD}

This section presents our approach to image denoising by learning a Convolutional Neural Network consisting of a series of Identity Mapping Modules. Each module is composed of a series of pre-activation units followed by convolution functions, with residual connections between them. 
The meta-structure of our Identity Enhanced Residual Denoising (IERD) network is explained in Section~\ref{sec:network_architecture} followed by the formulation of the learning objective in Section~\ref{sec:learning_obj}.

\subsection{Network Design}
\label{sec:network_architecture}
Residual learning has recently delivered state of the art results for object classification~(\cite{He15Residual,He2016IM}) and detection~(\cite{Lin16FPN}), while offers training stability. Inspired by the Residual Network variant with identity mapping~(\cite{He2016IM}), we adopt a modular design for our denoising network. The design consists of a series of Identity Mapping modules.

\subsubsection{Network Elements}

Figure~\ref{fig:net_architecture} depicts the entire architecture, where identity mapping modules are shown as blue blocks, which are, in turn, composed of basic ReLU and convolution layers. The output of each module is a summation of the identity function and the residual function. 
 
Three parameters govern the meta-level structure of the network: $\mathbf{M}$ is the number of identity modules, $\mathbf{L}$ is the number of pairs of pre-activation and convolution layers in each module, and $\mathbf{C}$ is the number of output channels, which we fixed across all the convolution layers. 
 
The high-level structure of the network can be viewed as a chain of identity modules, where the output of each module is fed directly into the succeeding one. Consequently, the output of this chain is fed to a final convolution layer to produce a tensor with the same number of channels as the input image. At this point, the final convolution layer directly predicts the noise component from a noisy image. The noisy image/patch is then added to the input to recover the noise-free image.  
 
The identity mapping modules are the building blocks of the network, which share the following structure. Each module consists of two branches: a residual branch and an identity mapping branch. The residual branch of each module contains a series of layers pairs, \ie a nonlinear pre-activation (typically ReLU) layer, followed by a convolution layer. Its primary responsibility is to learn a set of convolution filters to predict image noise. Besides, the identity mapping branch in each module allows the propagation of loss gradients in both directions without any bottleneck.  

\subsubsection{Justification of the network design}
Several previous image denoising works have adopted a fully convolutional network design, without any pooling mechanism~(\cite{zhang2017DnCNN,kim2016VDSR}). This is necessary in order to preserve the spatial resolution of the input tensor across different layers. We follow this design by using only non-linear activations and convolution layers across our network.

Furthermore, we aim to design the network in such a way where convolution layers neurons in the last layer of each identity mapping (IM)  module observe the full spatial receptive field in the first convolution layer. This design helps to learn to connect input neurons at all spatial locations to the output neurons, in much the same way as well-known non-local mean methods such as~(\cite{Dabov2007BM3D,Buades2005NLM}). Instead of using a unit dilation stride within each layer, we also experimented with dilated convolutions to increase the receptive fields of the convolution layers. By this design, we can reduce the depth of each IM module while the final layer's neurons can still observe the full input spatial extent. 

Pre-activation has been shown to offer the highest performance for classification when used together with identity mapping~(\cite{He2016IM}). In a similar fashion, our design employs ReLU before each convolution layer. This design differs from existing neural network architectures for denosing~(\cite{kim2016VDSR,lefkimmiatis2017NLNet}). The pre-activation helps training to converge more easily, while the identity function preserves the range of gradient magnitudes. Also, the resulting network generalizes better as compared to the post-activation alternative. This property enhances the denoising ability of our network.

\subsubsection{Formulation}
Now we formulate the prediction output of this network structure for a given input patch $\textbf{y}$. Let $\mathcal{W}$ denote the set of all the network parameters, which consists of the weights and biases of all constituting convolution layers. Specifically, we let $w_{m, l}$ denote both the kernel and bias parameters of the $l$-th convolution layer in the residual branch of the $m$-th module. 
 
Within such a branch, the intermediate output of the $l$-th ReLU-convolution pair and of the $m$-th module is a composition of two functions
\begin{equation}
\mathbf{z}_{m, l} = f(g(\mathbf{y}_{m, l}); w_{m, l}),
\end{equation}
where $f$ and $g$ are the notation for the convolution and the ReLU functions,  $\mathbf{z}_{m, l}$ is the output of the $l$-th ReLU-convolution pair of $m$-th module. 
By composing the series of ReLU-convolution pairs, we obtain the output of the $m$-th residual branch as
\begin{equation}
\begin{split}
\mathbf{r}_{m} = -\mathbf{z}_{m, 0} + f(g( \ldots f(g(\mathbf{y}_{m,0}; w_{m, 0})) \ldots ); w_{m, l}),
\end{split}
\end{equation}

where $\mathbf{z}_{m, 0}$ is the output of the first ReLU-convolution pair, and $\textbf{r}_m$ is the residual output of the corresponding module. Chaining all the identity mapping modules, we obtain the output as $\sum_{m=1}^{M} \textbf{r}_m$. Finally, the output of this chain is convolved with a final convolution layer with learnable parameters $w_{m+1}$ to  predict the noise component as 
$\text{IERD}(\mathbf{y}, \mathcal{W}) = f(\mathbf{y} + \sum_{m=1}^{M} \textbf{r}_m, w_{m+1})$.

\subsection{Learning to Denoise}
\label{sec:learning_obj}

Our network is trained on image patches or regions rather than at the entire image. A number of reasons drive this decision. 
Firstly, it offers a random sampling of a large number of training samples at different locations from various images. The random shuffling of training samples is well-known to be a useful technique to stabilize the training of deep neural networks. Therefore, it is preferable to batch training patches with a random, diverse mixture of local structures, patterns, shapes, and colors.
Secondly, there has been a success in approaches that learns image patch priors from external data for image denoising~(\cite{Zoran2011EPLL}).

From a set of noise-free training images, we randomly crop several training patches $\mathbf{x}_i, i=i, \ldots, N$ as the ground-truth. The noisy version of these patches is obtained by adding (Gaussian) noise to the ground truth training images. Let us denote the set of noisy patches corresponding to the former as $\mathbf{y}_i, i=i, \ldots, N$. With this setup, our image denoising network is aimed to reconstruct a patch $\mathbf{x}^*_i =\text{IERD}(\mathbf{y}_i, \mathcal{W})$ from the input patch $\mathbf{y}_i$. 

The learning objective is to minimize the following sum of squares of $\ell_2$-norms 
\begin{equation}
\mathcal{L} \triangleq \frac{1}{N}\sum_{i=1}^{N}  \Vert \text{IERD}(\mathbf{y}_i, \mathcal{W}) - \mathbf{x}_i \Vert^2.
\label{eq:objective}
\end{equation}

To train the proposed network on a large dataset, we minimize the objective function in Equation~\ref{eq:objective} on mini-batches of training examples. Training details for our experiments are described in Section \ref{label:training}.

\section{Experiments}

\subsection{Datasets}
\label{sec:Datasets}
We performed experimental validation on three widely used publicly available synthetically generated noisy datasets (in supplementary materials) and three real noisy image datasets described below.
\begin{itemize}
\item \textbf{DnD:} Recently, \cite{plotz2017benchmarking} proposed the Darmstadt Noise Dataset (DND) to benchmark the denoising algorithms. The dataset is composed of images with interesting and challenging structures. The size of each image is in Megapixels; therefore, each image is cropped at 20 locations of size 512 $\times$ 512 pixels yielding 1000 test crops. Only these test images are provided; there are no images for training or validation. 

\item \textbf{RNI15:}  RNI15 proposed by~\cite{lebrun2015NC} consists of 15 real noisy images. There are no ground-truth images available for this dataset.
\item \textbf{SIDD:} Smartphone Image Denoising Dataset (SIDD) proposed by \cite{abdelhamed2018SSID} is the largest collection of real-noisy images.  A total of 30k noisy images are gathered from ten different scenes under different lighting conditions via five smartphone cameras with their ground truth images. 
\end{itemize}

For evaluation purposes, we use the Peak Signal-to-Noise Ratio (PSNR) index as the error metric. We compare our proposed method with around 20+ state-of-the-art methods on the above six datasets. To ensure a fair comparison, we use the default setting provided by the respective authors. 

\begin{table}[!t]
\caption{Detailed architecture of an identity mapping module.}
\centering
\begin{tabular}{|l|c|c|c|c|c|c|c|}
\hline
&\multicolumn{6}{c|}{Identity Module Layers}\\ \cline{2-7}
Parameters  & 1$^{st}$  & 2$^{nd}$ & 3$^{rd}$ & 4$^{th}$ & 5$^{th}$ & 6$^{th}$\\ \hline 
Padding     & 1   		& 3 	   & 3        & 3        & 3        & 3\\
Dilation    & 1   		& 3 	   & 3        & 3        & 3        & 3\\ 
Kernel Size & 3   		& 3 	   & 3        & 3        & 3        & 3\\
Channels    & 64   		& 64 	   & 64       & 64       & 64       & 64\\ \hline
\end{tabular}
\label{table:Mapping_Module_architecture}
\end{table}

\subsection{Training Details}
\label{label:training}
The training input to our network is noisy, and noise-free patch pairs cropped randomly from the BSD400 dataset~(\cite{Martin2001BSD}) for synthetic denoising while for real noisy images, we use cropped patches from SSID~(\cite{abdelhamed2018SSID}), Poly(~\cite{xu2018real}), and RENOIR~(\cite{anaya2018renoir}). Note that there is no overlap between the training and evaluation datasets. We also augment the training data with horizontally and vertically flipped versions of the original patches and those rotated at an angle of $\frac{\pi n}{2}$, where $n=1,2,3$. The training patches are randomly cropped on the fly from the images of the mentioned datasets. 

We offer two strategies for handling different noise levels. The first one is to train a network for each specific noise level, and we call this model as \enquote{noise-specific} model. Alternatively, we train a single model for the any noise, and we refer to this model as a \enquote{noise-agnostic} model. At each update of training, we construct a batch of $32$ by randomly selecting noisy patches with different noise levels.  

We implement the denoising method in the PyTorch framework on two Tesla P100 GPUs and employ~\cite{KingmaB14}'s Adam optimization algorithm for training.
The initial learning rate was set to $10^{-4}$, and the momentum parameter was $0.9$. We scheduled the learning rate such that it is halved after every 10$^5$ iterations. We train our network from scratch by a random initialization of the convolution weights according to the method in~\cite{He15Rectifiers} and a regularization strength, \ie weight decay, of 10$^{-3}$.

\begin{table}[!t]
\caption{Denoising performance (in PSNR) on the BSD68 dataset~(\cite{Martin2001BSD}) for different sizes of training input patches for $\sigma_n = 25$, keeping all other parameters constant.}
\centering
\begin{tabular}{|c|c|c|c|c|c|}
\cline{1-6}
\multicolumn{6}{|c|}{Training patch size}\\ \hline
  20 	  & 30  	 & 40 		& 50 		&60  & 70\\ \hline 
  29.13  & 29.30 	 & 29.34   & 29.36    & 29.37 & 29.38\\ 
\hline
\end{tabular}
\label{table:Training_patch_size}
\end{table}
\subsection{Boosting Denoising Performance}
\label{sec:self_ensemble}
To boost the performance of the trained model, we use the late fusion/geometric transform strategy as adopted by~\cite{timofte2016seven}. During the evaluation, we perform eight types of augmentation (including identity) of the input noisy images $y$ as $y_i^t = \Gamma_i(y)$ where $i=1,\cdots,8$. From these geometrically transformed images, we estimate corresponding denoised images $\{\hat{x}_1^t, \hat{x}_2^t,\cdots, \hat{x}_8^t\}$, where $\hat{x}_i^t = \text{IERD}(\hat{y}_i^t,W)$  using our model. To generate the final denoised image $\hat{x}$, we perform the corresponding inverse geometric transform $\tilde{x}_i^{-t} = \Gamma_i^{-1}(\tilde{x}_i^t)$ and then take the average of the outputs as $\tilde{x} = \frac{1}{8}  \sum_{i=1}^{8} \tilde{x}_i^t $.
This strategy is beneficial as it saves training time and has a small number of parameters as compared to individually trained eight models. We also found empirically that this fusion method gives approximately the same performance as the models trained individually with geometric transform. The boosted version is denoted 

\subsection{Structure of Identity Modules}
The structure of the identity modules used in our experiments is depicted in Table~\ref{table:Mapping_Module_architecture}. Each module consists of a series of layers of \enquote{ReLU + Conv} pair. All the convolution layers have a kernel size of $3\times3$ and $64$ output channels. The kernel dilation and padding are the same in each layer and vary between $1$ and $3$. The skip connection connects the output of the first pair of \enquote{ReLU + Conv} to the last \enquote{Conv} as shown in figure~\ref{fig:net_architecture}

\subsection{Ablation Studies}

\subsubsection{Influence of the patch size} 
In this section, we show the role of the patch size and its influence on the denoising performance.
Table~\ref{table:Training_patch_size} shows the average PSNR on BSD68~(\cite{roth2009fields}) for $\sigma_n=25$ with respect to the increase in size of the training patch. It is obvious that there is a marginal improvement in PSNR as the patch size increases. The main reason for this phenomenon is the size of the receptive field, with a larger patch size network learns more contextual information, hence able to predict local details better.

\begin{table}
\caption{The average PSNR of the denoised images for the BSD68 dataset, with respect to different number of modules $\textbf{M}$. The higher the number of modules, the higher is the accuracy.}
\centering
\begin{tabular}{|c|c|c|c|c|}
\cline{1-5}
\multicolumn{5}{|c|}{Number of modules}                \\ \hline
2   	&3     &4    	&6 		 &8   \\ \hline 
29.28	&29.34 &29.34 	&29.35   &29.36	 \\ \hline
\end{tabular}
\label{table:No_of_cnn_modules}
\end{table}

\begin{table}
\caption{Denoising performance for different network settings to dissect the relationship between kernel dilation, number of layers and receptive field.}
\centering
\begin{tabular}{|c|c|c|c|}
\cline{1-4}
No of layers     & 18   	& 9  	    & 6      	\\ \hline
Kernel dilation  & 1   		& 2   	    & 3      	\\ \hline 
      	  		 & 29.34      & 29.34  	& 29.34  	\\ 
\hline
\end{tabular}
\label{table:layers_kernel_dilation_patchsize}
\end{table}

\subsubsection{Number of modules}
We show the effect of the number of modules on denoising results. As mentioned earlier, each module $\textbf{M}$ consists of six convolution layers, by increasing the number of modules, we are making our network deeper. In this settings, all parameters are constant, except the number of modules, as shown in Table~\ref{table:No_of_cnn_modules}. It is clear from the results that making the network deeper increases the average PSNR.  However, since fast restoration is desired, we prefer a small network of three modules \ie $\textbf{M}=3$, which still achieves better performance than competing methods.

\subsubsection{Kernel dilation and number of layers}
It has been shown that the performance of some networks can be improved either by increasing the depth of the network or by using large convolution filter size to capture the context information~(\cite{zhang2017IRCNN,zhang2017DnCNN}). This helps the restoration of noisy structures in the image. The usage of traditional $3\times3$ filters is popular in deeper networks. However, there is a tradeoff between the number of layers and the size of the dilated filters without effecting denoising results. In Table~\ref{table:layers_kernel_dilation_patchsize}, we present three experimental settings to show the tradeoff between the dilated filter size and the depth of the network. In the first experiment, as shown in the first column of Table~\ref{table:layers_kernel_dilation_patchsize}, we use a traditional filter of size $3\times3$ and depth of 18 to cover the receptive field of training patch.

\begin{table}[tb]
\caption{PSNR reported on the BSD68 dataset for $\sigma_n = 25$ when different features are added to the baseline (first row).}
\centering
\begin{tabular}{|c|c|c|c|c|}\hline
Dilation & Identity & Boosting & PSNR\\ 
\hline
 	   &   &   & 29.24 \\ 
	 \checkmark &   &   & 29.23 \\
	   & \checkmark &   & 29.28 \\
	 \checkmark & \checkmark &   & 29.32 \\
	 \checkmark & \checkmark & \checkmark & 29.34 \\
\hline
\end{tabular}
\label{table:feature_contributions}
\end{table}

\begin{figure*}
\begin{center}
\begin{tabular}{c@{ } c@{ } c@{ } c@{ } c}
\includegraphics[width=.19\textwidth]{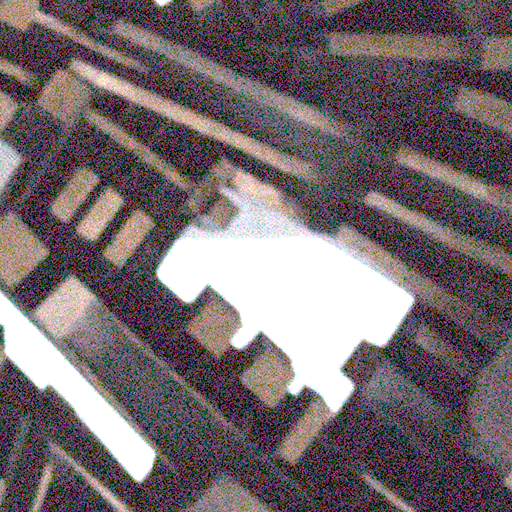}&
\includegraphics[width=.19\textwidth]{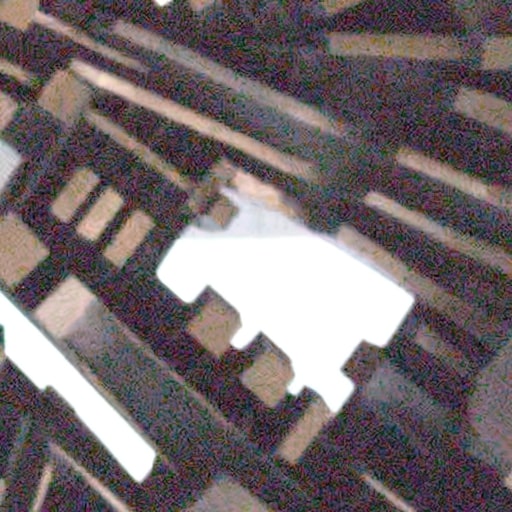}&
\includegraphics[width=.19\textwidth]{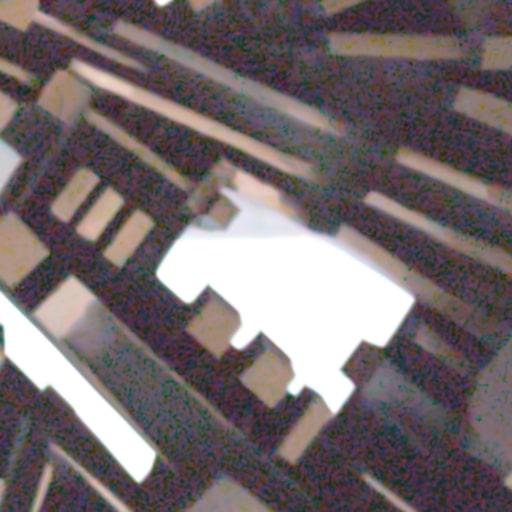}&
\includegraphics[width=.19\textwidth]{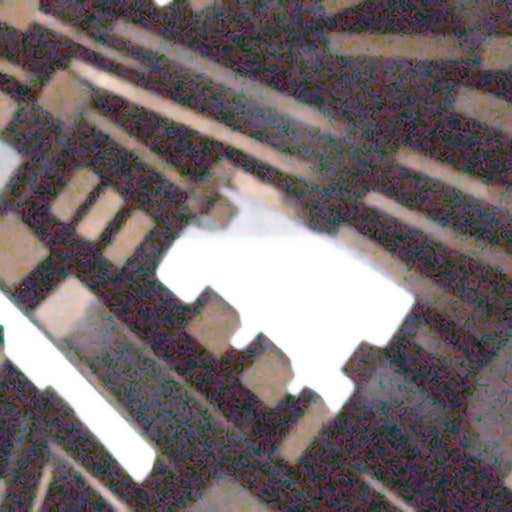}&
\includegraphics[width=.19\textwidth]{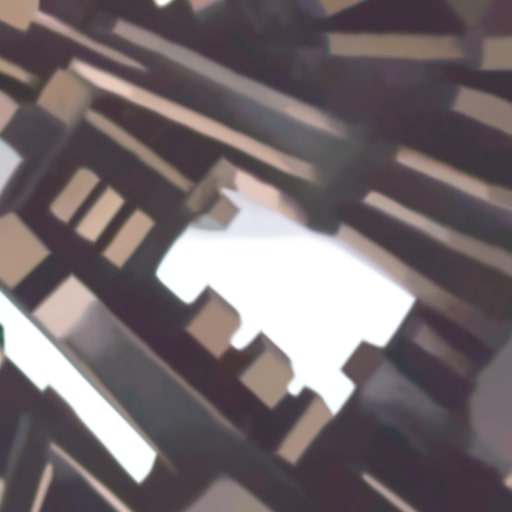}\\
   &23.95dB  & 25.63dB &27.28 & 32.97dB  \\

Noisy & CBM3D  & WNNM & TNRD &  TWSC \\

\includegraphics[width=.19\textwidth]{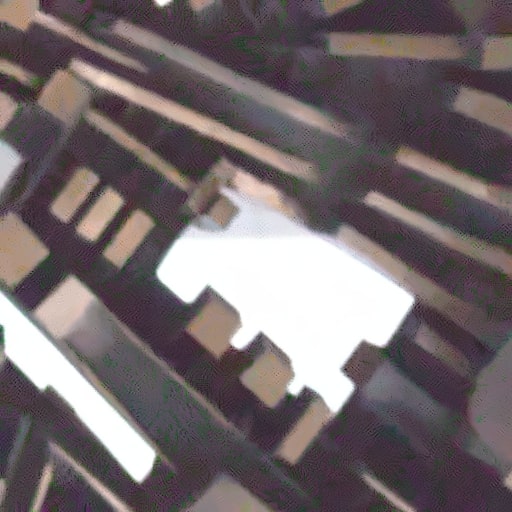}&
\includegraphics[width=.19\textwidth]{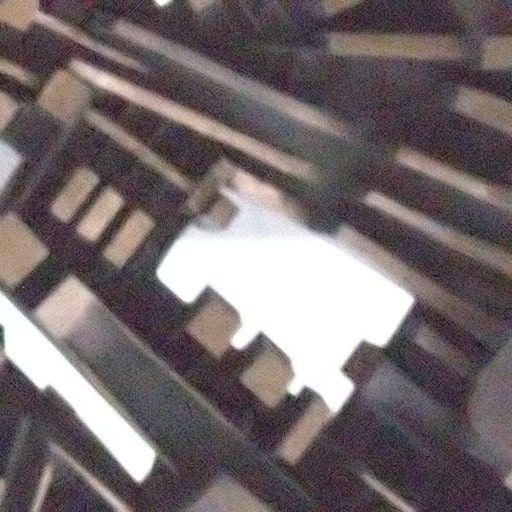}&
\includegraphics[width=.19\textwidth]{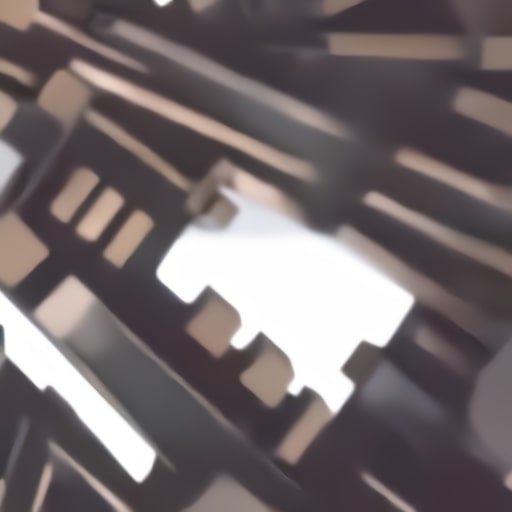}&
\includegraphics[width=.19\textwidth]{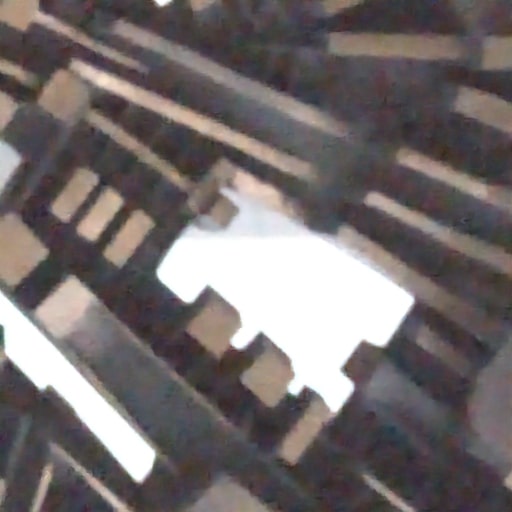}&
\includegraphics[width=.19\textwidth]{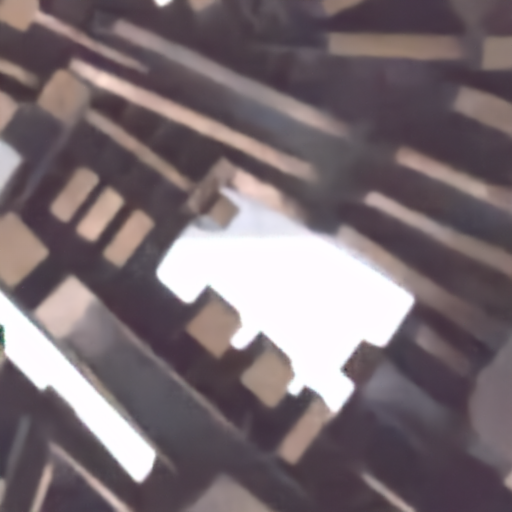}\\
28.32dB & 27.28dB & 32.14dB & 31.40dB & \textbf{33.79dB}\\
NC& NI  &FFDNet       & CBDNet & IERD (Ours)\\
\end{tabular}
\end{center}
\caption{Comparison of our method against the state-of-the-art algorithms on real images containing Gaussian noise from Darmstadt Noise Dataset (DND) benchmark for different denoising algorithms. Difference can be better viewed in magnified view.}
\label{fig:DnD2}
\end{figure*}



In the next experiment, we keep the size of the filter the same but enlarge the filter using a dilation factor of two. Although this increases the size of the filter to $5\times5$; however, still having only nine non-zero entries similar to the above experiment, and it can be interpreted as a sparse filter. Therefore, the receptive field of the training patch can now be covered by nine non-linear mapping layers, contrary to the 18-layers depth per module. Similarly, by expanding the filter by dilation of three would result in the depth of each module to be six. As in Table~\ref{table:layers_kernel_dilation_patchsize}, all three trained models result in similar denoising performance, with the apparent advantage of the shallow network being the most efficient. The number of parameters reduced from 1954k to 663k; similarly, the memory usage for one input patch is reduced from 22MB to 6.5MB.

\subsubsection{Network structure Analysis} 
In Table~\ref{table:feature_contributions}, we show the performance on the BSD68 dataset when adding different features, including a kernel dilation of three across all convolution layers, identity skip connection, or boosting via geometric transformation to the DnCNN baseline which is reported in the first row.
The improvement over DnCNN is observed with the introduction of identity skip connections. Applying a dilation of three over 17 or 19 convolutional layers of DnCNN (row 2) does not appear to be effective. However, using dilated convolution in a short chain of six layers, such as row 3, improves the performance further. In Table~\ref{table:feature_contributions}, PSNR is $29.32$ dB without boosting and $29.34$ dB (last row) if we average the output from eight transformed images.

\begin{table}[tp]
\caption{Mean PSNR and SSIM of the denoising methods evaluated on the real images dataset by \cite{plotz2017benchmarking}.}
\centering
\resizebox{\columnwidth}{!}{%
\begin{tabular}{l||ccc}
\hline \hline
Method                              &Blind/Non-blind &PSNR &SSIM \\ \hline \hline
CDnCNNB~(\cite{zhang2017DnCNN})  	&Blind 		&32.43 &0.7900\\
EPLL~(\cite{Zoran2011EPLL}) 		&Non-blind 	&33.51 &0.8244\\
TNRD~(\cite{chen2017TNRD})  		&Non-blind 	&33.65 &0.8306\\
NCSR~(\cite{dong2012NCSR})  		&Non-blind 	&34.05 &0.8351\\
MLP~(\cite{Burger2012MLP})  		&Non-blind 	&34.23 &0.8331\\
FFDNet~(\cite{zhang2018ffdnet})  	&Non-blind 	&34.40 &0.8474\\
BM3D~(\cite{Dabov2007BM3D})  		&Non-blind 	&34.51 &0.8507\\
FoE~(\cite{roth2009fields})  		&Non-blind 	&34.62 &0.8845\\
WNNM~(\cite{Gu2014WNN})  		    &Non-blind 	&34.67 &0.8646\\
NC~(\cite{lebrun2015NC})	        &Blind	    &35.43	&0.8841	\\
NI~(\cite{NeatI})	    	        &Blind	    &35.11  &0.8778 \\
KSVD~(\cite{aharon2006ksvd})  		&Non-blind 	&36.49 &0.8978\\
MCWNNM~(\cite{xu2017MCW})  		    &Non-blind 	&37.38 &0.9294\\
TWSC~(\cite{xu2018TWSC})            &Non-blind	&37.96 &0.9416\\
FFDNet+~(\cite{zhang2018ffdnet})	&Non-blind  &37.61 &0.9415\\
CBDNet~(\cite{guo2018CBDnet})		&Blind 		&38.06 &0.9421\\
IERD (Ours)		                    &Blind 		&\textbf{39.20} & \textbf{0.9524}\\
RIDNET~\cite{anwar2019real}         &Blind      &\textbf{39.25} & \textbf{0.9528}\\
IERD+ (Ours)	                    &Blind 		&\textbf{39.30} & \textbf{0.9531}\\ 
\hline \hline
\end{tabular}
}
\label{table:DnD_benchmark}
\vspace*{-5mm}
\end{table}

\subsection{Real-world images}

So far, state-of-the-art denoising methods, such as DnCNN~(\cite{zhang2017DnCNN}), IrCNN~(\cite{zhang2017IRCNN})  and BM3D~(\cite{Dabov2007BM3D}) \etc usually have been evaluated on classical images, and the BSD68 dataset  but their performance is limited on real image datasets. Furthermore, real image denoising is becoming popular; hence, we compare our method against recent state-of-the-art~\cite{anwar2019real,guo2018CBDnet,zhang2018ffdnet} algorithms.

\subsubsection{Darmstadt Noise Dataset}

\begin{figure*}[!t]
\begin{center}
\begin{tabular}{c@{ } c@{ } c@{ } c@{ } c}  

\includegraphics[trim={4cm 2cm  4cm  2cm},clip,width=.18\textwidth]{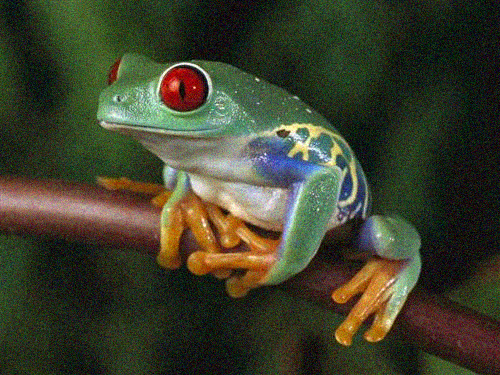}&  
\includegraphics[trim={4cm 2cm  4cm  2cm},clip,width=.18\textwidth]{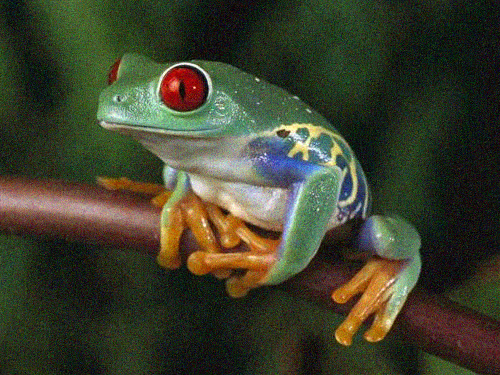}&  
\includegraphics[trim={4cm 2cm  4cm  2cm},clip,width=.18\textwidth]{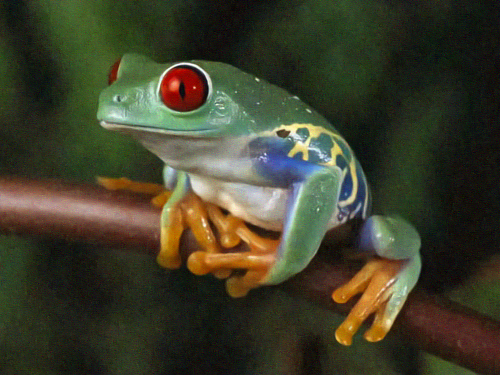}&
\includegraphics[trim={4cm 2cm  4cm  2cm},clip,width=.18\textwidth]{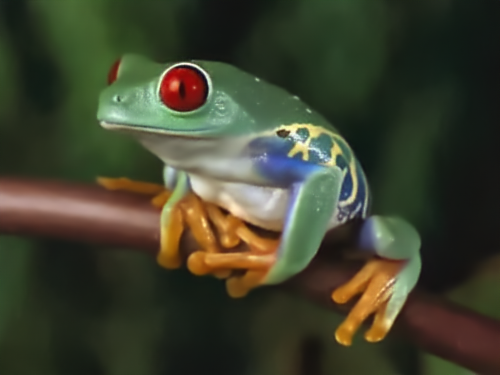}&
\includegraphics[trim={4cm 2cm  4cm  2cm},clip,width=.18\textwidth]{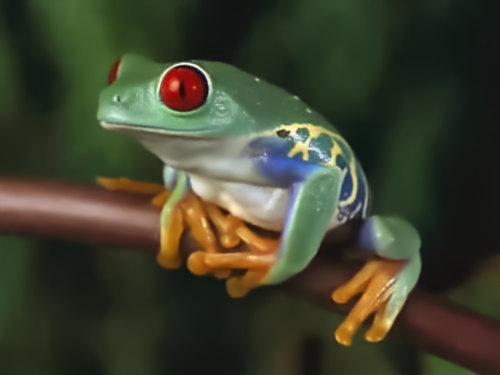}\\
Input   &   IRCNN & CBDNet& IERD (Ours) & IERD+ (Ours)\\

\includegraphics[trim={1.5cm 1.5cm  1.5cm  1cm },clip,width=.18\textwidth]{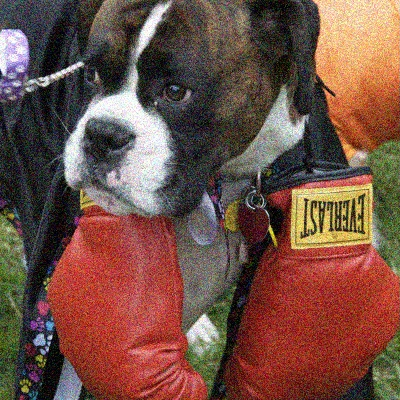}&
\includegraphics[trim={1.5cm 1.5cm  1.5cm  1cm },clip,width=.18\textwidth]{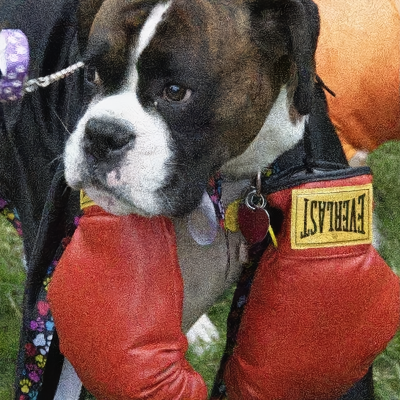}&  
\includegraphics[trim={1.5cm 1.5cm  1.5cm  1cm },clip,width=.18\textwidth]{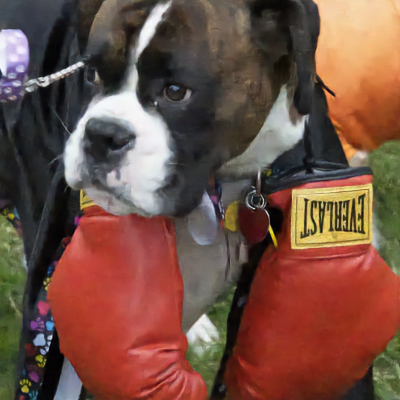}&  
\includegraphics[trim={1.5cm 1.5cm  1.5cm  1cm },clip,width=.18\textwidth]{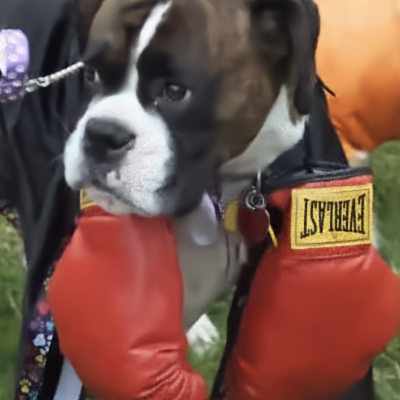}&
\includegraphics[trim={1.5cm 1.5cm  1.5cm  1cm },clip,width=.18\textwidth]{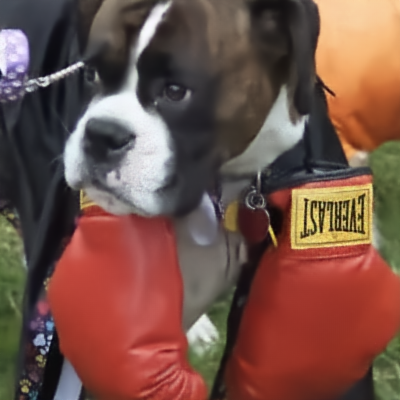}\\
Input   &   FFDNet & CBDNet& IERD (Ours) & IERD+ (Ours)\\

\end{tabular}
\end{center}
\caption {Sample visual examples from RNI15~(\cite{lebrun2015NC}). Our method annihilates the noise and preserves the essential details while the competing methods fail to deliver satisfactory results \ie unable to remove noise. Best viewed on high-resolution display.} 
\label{fig:RNI15_results}
\vspace*{-4mm}
\begin{center}
\begin{tabular}{c@{ } c@{ } c@{ } c@{ } c@{ } c@{ } c@{ } c}  
\includegraphics[width=.12\textwidth]{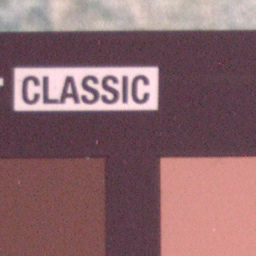}&  
\includegraphics[width=.12\textwidth]{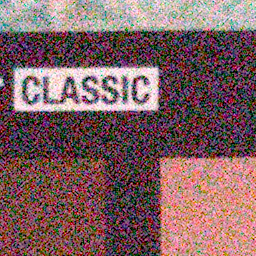}&  
\includegraphics[width=.12\textwidth]{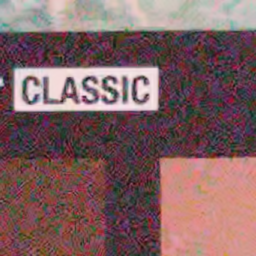}&  
\includegraphics[width=.12\textwidth]{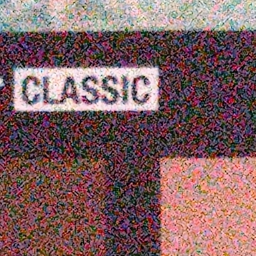}&  
\includegraphics[width=.12\textwidth]{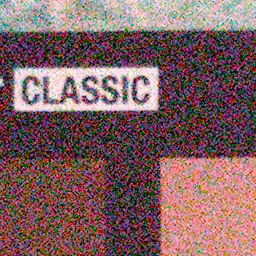}&  
\includegraphics[width=.12\textwidth]{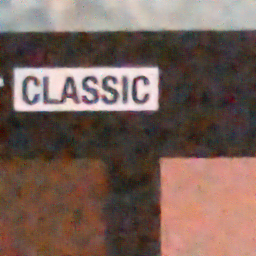}&
\includegraphics[width=.12\textwidth]{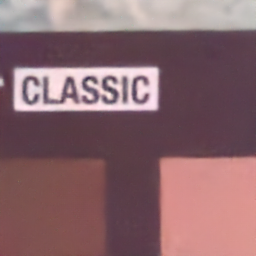}&
\includegraphics[width=.12\textwidth]{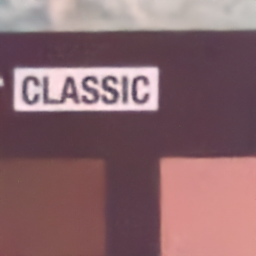}\\

\includegraphics[width=.12\textwidth]{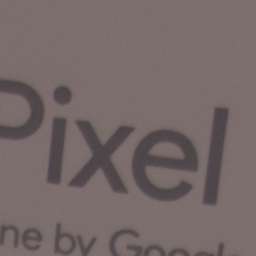}&  
\includegraphics[width=.12\textwidth]{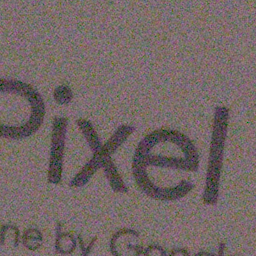}&  
\includegraphics[width=.12\textwidth]{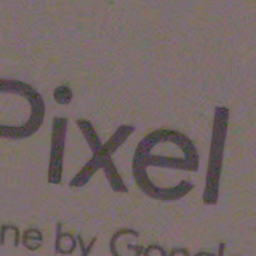}&  
\includegraphics[width=.12\textwidth]{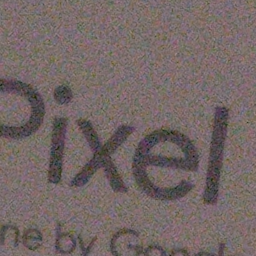}&  
\includegraphics[width=.12\textwidth]{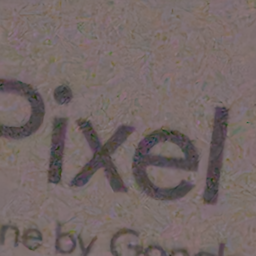}&  
\includegraphics[width=.12\textwidth]{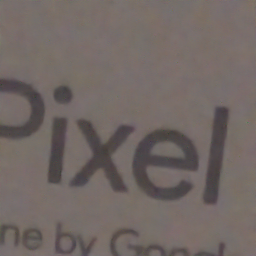}&
\includegraphics[width=.12\textwidth]{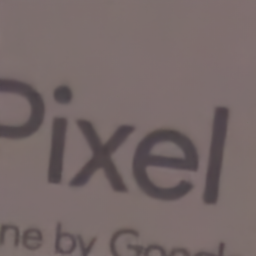}&
\includegraphics[width=.12\textwidth]{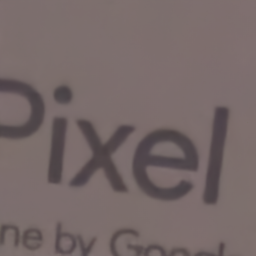}\\
GT   &  Noisy & CBM3D & DnCNN & FFDNet & CBDNet & IERD (Ours) & IERD+ (Ours)\\
\end{tabular}
\end{center}
\caption {A few challenging examples from SSID dataset~(\cite{abdelhamed2018SSID}). Our
method can restore true colors and remove noise.} 
\label{fig:SSID_results}
\vspace*{-4mm}
\end{figure*}

We visually compare our method with a few recent algorithms, as shown on several samples from \cite{plotz2017benchmarking} in Figure~\ref{fig:DnD2}. It can be observed that synthetic denoiser such as CBM3D~(\cite{Dabov2007BM3D}), DnCNN~(\cite{zhang2017DnCNN}) \etc, and real image denoisers such as CBDNet~(\cite{guo2018CBDnet}) and FFDNet~(\cite{zhang2018ffdnet}), are unable to remove the noise from the images. On the other hand, it can be seen that our method eliminates noise and preserve the structures.

The quantitative results in PSNR and SSIM averaged over all the images for real-world DnD is presented in Table~\ref{table:DnD_benchmark}. Our method is the best performer, followed by CBDNet. Our method is also able to improve significantly on NI~(\cite{NeatI}), a software which is part of coral draw and photoshop. It is to be noted that our method does not require to know the noise level in advance, like \cite{Dabov2007BM3D}'s BM3D and does not require to estimate it separately, like~\cite{guo2018CBDnet}'s CBDNet. 

\subsubsection{RNI15} The ground-truth images for RNI15~(\cite{lebrun2015NC}) are not publicly available; therefore, we present the visual comparison only in Figure~\ref{fig:RNI15_results}. In the first example, we can see that there are artifacts on the face for the output of FFDNet~(\cite{zhang2018ffdnet}) and CBDNet~(\cite{guo2018CBDnet}) while our method is able to remove the noise without introducing any artifacts. In the second example (given in the second row), our method smooths out the noise and can produce crisp edges while the competing method fails to produce any results without noise. The noise structures are very prominent in the second image near the eyes, as well as the gloves. This shows the robustness of our method against challenging images.

\subsubsection{SSID}
We utilize the SIDD real noise dataset~(\cite{abdelhamed2018SSID}) as the final dataset for comparison. Table~\ref{table:SSID} shows the average PSNR on the validation dataset where our method improves upon FFDNet~(\cite{zhang2018ffdnet}) and CBDNet~(\cite{guo2018CBDnet}) with a margin of 9.62dB and 8.04dB. Next, we show the sample visual denoise images from SIDD for various competing algorithms in Figure~\ref{fig:SSID_results}. Our results are resembling the ground-truth image colors while the previous state-of-the-art images produce color casts and artificial colors.

\begin{table}
\caption{The quantitative results (in PSNR (dB)) for the SSID dataset~(\cite{abdelhamed2018SSID}).}
\centering
\resizebox{\columnwidth}{!}{%
\begin{tabular}{c|c|c|c|c|c}
\hline\hline
\multicolumn{6}{c}{Methods} \\ 
BM3D     &  DnCNN   & FFDNet	 & CBDNet&RIDNet    & IERD+	  \\\hline 
30.88    &  26.21   & 29.20   & 30.78     &\textbf{38.71} & \textbf{38.82} \\ \hline \hline
\end{tabular}}
\label{table:SSID}
\vspace*{-5mm}
\end{table}
\section{Conclusions}
\label{sec:Conclusion}
To sum up, we employ residual learning and identity mapping to predict the denoised image using a three-module and six-layer deep network of 19 weight layers with dilated convolutional filters without batch normalization. Our choice of network is based on the ablation studies performed in the experimental section of this paper. 

This is the first modular framework to predict the denoised output without any dependency on the pre- or post-processing. Our proposed network removes the potentially authentic image structures while allowing the noisy observations to go through its layers, and learns the noise patterns to estimate the clean image.

On real images, we have shown that our method provides visually pleasing results and a gain of about $\textbf{1.2}$dB on Darmstadt Noise Dataset, $\textbf{9.62}$dB on smartphone image denoising dataset (SIDD) in terms of PSNR. The real images appear less grainy after passing through our proposed network and preserving fine image structures. Furthermore, competitive denoising algorithms either require information about the noise in advance or estimate it in a disjoint stage while, on the contrary, our network does not require any information about the noise present in the images. 

In the future, we aim to generalize our denoising network to other image restoration and enhancement tasks such as deblurring, color correction, JPEG artifact removal, rain removal, dehazing, and super-resolution \etc

\clearpage
{\small
\bibliographystyle{ieee_fullname}
\bibliography{egbib}
}

\end{document}